\documentclass[letterpaper, 10 pt, conference]{ieeeconf}  

\IEEEoverridecommandlockouts                              

\usepackage{graphics} 
\usepackage{epsfig} 
\usepackage{mathptmx} 
\usepackage{times} 
\usepackage{amsmath} 
\usepackage{amssymb}  
\usepackage{bm}
\usepackage{hyperref}

\usepackage{algorithm}
\usepackage{algorithmic}
\usepackage{bbm}
\usepackage{multirow}
\usepackage{caption}
\usepackage{color}
\title{\LARGE \bf
Diff-Net: Image Feature Difference based High-Definition Map Change Detection for Autonomous Driving
}

\author{Lei He$^{1}$, Shengjie Jiang$^{2}$, Xiaoqing Liang$^{2}$, Ning Wang$^{2}$ and Shiyu Song$^{1*}$
\thanks{$^{1}$The authors are with Baidu Autonomous Driving Technology Department, \texttt{\{helei07, songshiyu\}@baidu.com.}}%
\thanks{$^{2}$Work done during internship at Baidu, \texttt{2104190011@cugb.edu.cn}, \texttt{xiaoqing.liang@nlpr.ia.ac.cn}, \texttt{20181222016@nuist.edu.cn}.}
\thanks{$^{*}$Author to whom correspondence should be addressed, E-mail: {\tt\small \href{mailto:songshiyu@baidu.com}{songshiyu@baidu.com}}.}%
}

\begin{document}

\maketitle
\thispagestyle{empty}
\pagestyle{empty}

\begin{abstract}

Up-to-date High-Definition (HD) maps are essential for self-driving cars. To achieve constantly updated HD maps, we present a deep neural network (DNN), Diff-Net, to detect changes in them. Compared to traditional methods based on object detectors, the essential design in our work is a parallel feature difference calculation structure that infers map changes by comparing features extracted from the camera and rasterized images. To generate these rasterized images, we project map elements onto images in the camera view, yielding meaningful map representations that can be consumed by a DNN accordingly. As we formulate the change detection task as an object detection problem, we leverage the anchor-based structure that predicts bounding boxes with different change status categories. To the best of our knowledge, the proposed method is the first end-to-end network that tackles the high-definition map change detection task, yielding a single stage solution. Furthermore, rather than relying on single frame input, we introduce a spatio-temporal fusion module that fuses features from history frames into the current, thus improving the overall performance. Finally, we comprehensively validate our method's effectiveness using freshly collected datasets. Results demonstrate that our Diff-Net achieves better performance than the baseline methods and is ready to be integrated into a map production pipeline maintaining an up-to-date HD map.

\end{abstract}

\section{INTRODUCTION}

Autonomous driving has moved from the realm of science fiction to a practical possibility during the past twenty years.
Despite many years of research and development, building reliable solutions that can handle the complexity of the real world remains an open problem.
Modern self-driving vehicles primarily rely on detailed pre-built maps, the high-definition (HD) map, which typically contains rich information of the environment, such as topology and location of lanes, crosswalks, traffic lights, and intersections.

They are a great source of prior knowledge and must be maintained adequately by constantly updating them to reflect up-to-date changes in the real world.
These changes typically include recently installed or removed traffic signals, portable traffic signals that just moved to new positions, or human errors
during map production.
In this work, we address the HD map update problem by building an end-to-end learning-based network that detects changes in the HD map, helping our self-driving vehicles access the latest environmental information.

As we aim to detect changes in an HD map, a conventional approach is to leverage object detection algorithms together with necessary association and difference calculation algorithms.
In this way, it derives map element changes given existing map elements and object detection results from online camera imagery.
The entire processing pipeline involves multiple steps, such as object detection, element association, and difference calculation.
However, the apparent problem is that each step above has its optimization objective, making the entire change detection pipeline fail to achieve an overall optimal solution.
For example, an object detector typically involves thresholding detection confidence scores and running Non-Maximum Suppression (NMS) to trade-off precision against recall.
Falsely detected objects in the early step will quickly lead to change detection failures eventually.
Meanwhile, the traditional approach ignores important prior information from the HD map.

In this work, we propose an end-to-end learning-based method to detect map changes directly.
More specifically, we use a deep neural network (DNN) to detect missing or redundant elements in an HD map.
To incorporate the prior information in HD maps, we project map elements onto images and rasterize them from the camera's perspective.
Both the rasterized and online camera images are furnished into the DNN as input.
We then calculate the differences between extracted features from both sources in different scales.
These feature differences are propagated, fused, and decoded, finally yielding an end-to-end HD map change detection (HMCD) network.
Moreover, since the changes are consistent in a group of consecutive frames over time, it is worth mentioning that we introduce a spatio-temporal feature fusion module to improve its performance further.
To fully validate the designs in our proposed method, we construct a large-scale dataset that includes abundant synthesized and real HD map change cases.
The synthesized cases help us overcome the vital issue that HD map changes are low probability events in practice, thus allowing us to accomplish network training and performance evaluation.

\section{Related Work}
Although HD maps have become an indispensable module in an autonomous driving system in recent years, relatively few attempts specifically focus on the HD map change detection (HMCD) problem in the academic community. Pannen et al.~\cite{pannen2019hd} propose a crowd-based method that combines particle filter and boosted classifier to infer the probability of HD map changes. Heo et al.~\cite{heo2020hd} adopt an encoder-decoder architecture driven by adversarial learning, achieving a pixel-level HD map change detector in the camera view.

The most related task probably is the scene change detection~\cite{radke2005image,taneja2015geometric,sakurada2015change,qin20163d,palazzolo2018fast,lei2020hierarchical}, of which the solutions can be mainly divided into three categories.
The first category leverages 3D-to-3D comparisons~\cite{golparvar2011monitoring} between a pre-built 3D CAD model and a reconstructed one built by classic multi-view stereo (MVS) methods~\cite{agarwal2011building,furukawa2015multi}, known to be high time-consuming methods and only applicable for offline applications.
The second approach~\cite{pollard2007change,ulusoy2014image,taneja2011image,qin20143d,palazzolo2018fast} is to infer the changes of the scene by comparing newly acquired images against the original 3D model.
In particular, the probability of changes can be estimated by comparing the voxel color of a 3D voxel-based model against the color of the corresponding image pixels~\cite{pollard2007change,ulusoy2014image}.
A relevant alternative~\cite{taneja2011image,qin20143d,palazzolo2018fast} is to identify changes by re-projecting a new image onto an old one with the help of the given 3D model and compare the inconsistencies.
The third category~\cite{radke2005image,sakurada2013detecting,eden2008using,sakurada2015change,zhan2017change,guo2018learning,alcantarilla2018street,sakurada2020weakly,lei2020hierarchical} adopts 2D-to-2D comparisons between images representing old states and current states of a scene.
A prior 2D-to-2D image registration step is required.

Besides detecting changes in a scene, our HD map change detection task identifies changed elements in the HD map together with the types of changes.
A straightforward method is to recognize map elements in images using a standard object detector, project map elements onto the images, associate the projections with the detections, and finally obtain the corresponding changes through a cross-comparison procedure.
Object detection is a classic problem in computer vision.
The solutions can be mainly divided into two categories, namely two-stage~\cite{ren2016faster,lin2017feature,he2017mask} and one-stage~\cite{redmon2018yolov3,law2018cornernet,liu2016ssd} methods.

This work introduces an image feature difference-based HD map change detection method that infers the map changes by adopting the anchor-based one-stage detection method, YOLOv3~\cite{redmon2018yolov3}, as its detection head.

\section{Problem Formulation}
\label{section:problem_formulation}

The HD map change detection (HMCD) task is formulated similar to an object detection problem.
The goal is to recognize map change instances of a predefined set of object classes (e.g., traffic lights, signs), describe the locations of detected objects in the image using 2D bounding boxes, and assign correct change categories for them, including $to\_add$, $to\_del$, and $correct$.
As their names imply, objects with $to\_add$ attributes are the ones missed, $to\_del$ are the ones that should be removed, and $correct$ are the correct ones in the HD map, respectively.
Portable traffic signals are special cases as they are treated as a pair of $to\_del$ and $to\_add$ bounding boxes illustrating both the old and new locations of the traffic signals.
In particular, we focus on the change detection task of traffic signals in this work.
Note that our proposed method can be extended to other objects in regular shapes, while the change detection problem of irregularly shaped ones is beyond the scope of this work.

Formally, for an online HMCD method that works with a single image as input, the problem can be formulated as:
\begin{equation}\label{eq:diff_formulation}
	\bm{D}_{k} = f_{\theta}(M, I_{k}, T_{k}, K),
\end{equation}
where $I_{k}$ is the $k$-th image frame in a video stream, $T_k$ is a global camera pose typically estimated by a localization system in a self-driving car, $K$ is the camera intrinsic calibration matrix, $M$ is the HD map, and $\bm{D}_k$ is a set of 2D bounding boxes with corresponding change categories predicted by our HMCD predictor $f_{\theta}$ with a set of learnable parameters $\theta$.

\section{Method}

\begin{figure*}[htb]
	\begin{center}
		\includegraphics[width=0.88\linewidth]{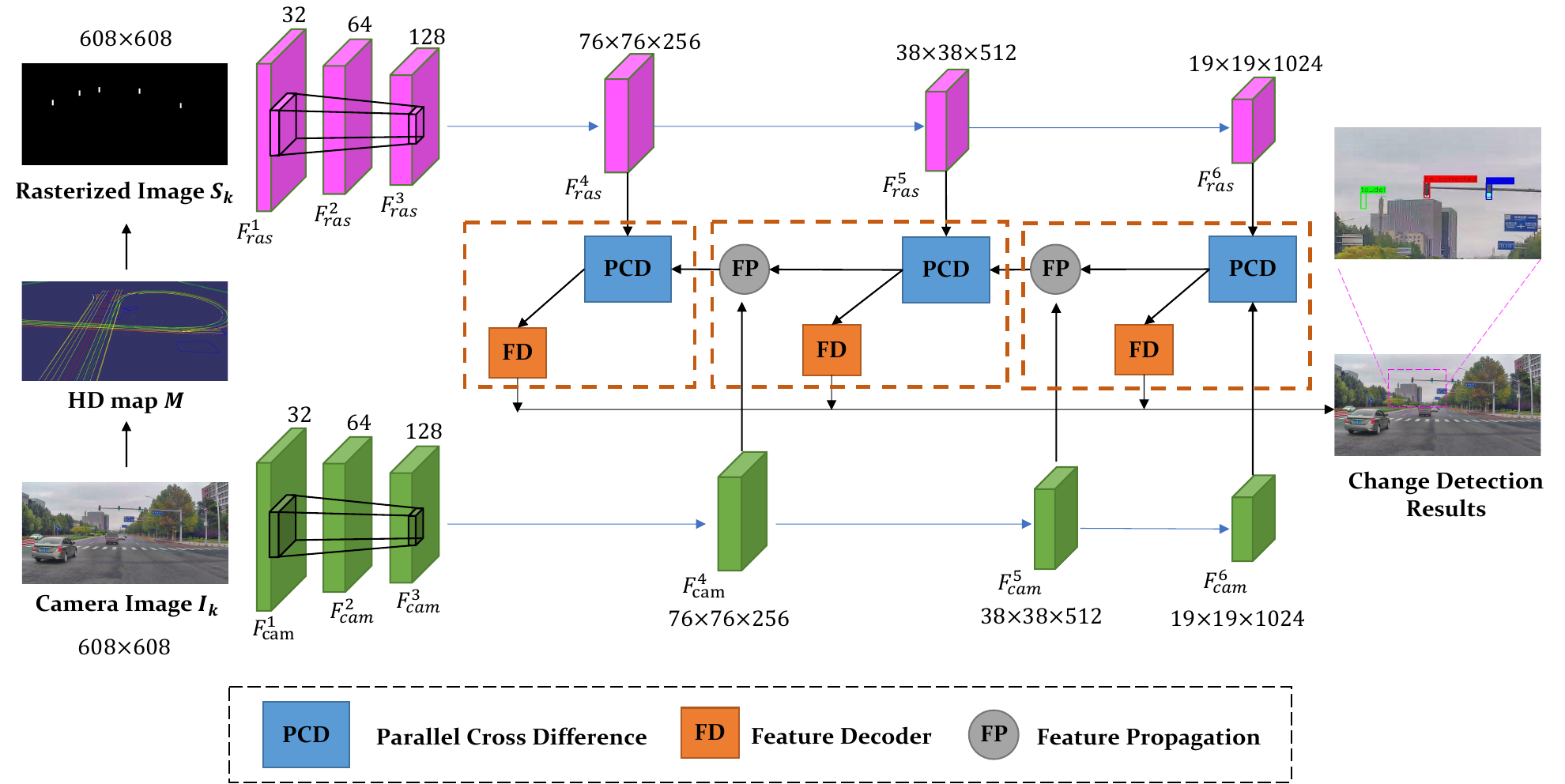}
	\end{center}
	\caption{The overall architecture of the Diff-Net involving three main modules: (a) rasterized images including projected map elements used as one of the inputs to CNN; (b) CNN based pyramid feature extraction layers for two parallel model inputs (pink and green); (c) a series of feature operators that involve feature propagation (FP), parallel cross difference (PCD) calculation, and feature decoding (FD).
	We downsample the original images from $1920 \times 1080$ to $608 \times 608$ as the network input.
	$I_{k}$ is the $k$-th camera image.
	$M$ is the HD map.
	$S_k$ is the rasterized image embedded with map labels.
	$F_{ras}^{i}$ and $F_{cam}^{i}$ denote the extracted features from rasterized and camera images at $i$-th scale level, respectively.
	In addition, 32, 64, and 128 are the numbers of feature channels in the first, second, and third scale levels.
	The blue, green, and red masks in the top-right image represent the ground truth detection results in $correct$, $to\_del$, and $to\_add$ change categories, respectively.}
	\label{fig:diff-net}
\end{figure*}

The overall architecture of the proposed Diff-Net is shown in Figure~\ref{fig:diff-net}.
Besides using the original camera image as our model input, we first construct a rasterized image by projecting map elements onto it from the camera perspective.
Then, pyramid features in different resolutions are extracted by two parallel CNN-based backbones from both the rasterized and camera images.
The key design of our work is to infer map changes by having a series of feature operators that propagate, fuse, differentiate, and decode these features.
Meanwhile, we leverage anchor-based object detection techniques, finally inferring map changes from decoded features.
The following sections describe them in detail.

\subsection{Model Input}
The fact that HD maps and camera images are data in different modalities poses considerable challenges in building a meaningful map data representation that a deep neural network can consume as an input.
Inspired by recent planning~\cite{bansal2019chauffeurnet,buhler2020driving,hecker2020learning}, prediction~\cite{casas2018intentnet,chai20a,Djuric_2020_WACV} or change detection~\cite{heo2020hd} works, we propose to construct an image from the camera perspective and rasterize it by projecting map elements onto it.
Specifically, given a global pose (position and orientation) of the camera, we first query map elements within a region of interest (ROI) in the HD map.
Then, they are projected onto the canvas from the camera perspective, with the projected area filled with a homochromatic color for the same type of objects in the HD map.
This yields a binary image, if we only focus on a single object category, such as traffic lights, as shown in an example in the upper left corner of Figure~\ref{fig:diff-net}.
Furthermore, it is worth noting that our proposed method can be generalized to multi-category objects by rendering them with different colors.
And it has been found that different color selections do not produce a significant effect when images are rasterized similarly in other tasks \cite{Djuric_2020_WACV}.

\subsection{Feature encoding and Difference}
Given both the rasterized images embedded with local map information and online camera images, we aim to extract meaningful features from them, yielding two parallel feature extraction pipelines in our implementation.
They are shown in pink and green colors in Figure~\ref{fig:diff-net}.
We adopt a shallow 11-layers CNN to increase its receptive field for the rasterized images, where the convolution strides of the 3rd, 6th, and 8-11th layers are 2, and others are 1.
The size of the convolution kernel is $3\times3$, and the number of channels is shown in Figure~\ref{fig:diff-net}.
It is verified to be sufficient for feature extraction from these relatively clean images.
For feature extraction of online camera images, we use DarkNet-53 from YOLOv3~\cite{redmon2018yolov3} because of its well-balanced accuracy and inference speed.

\begin{figure*}[htb]
	\begin{center}
		\includegraphics[width=0.88\linewidth]{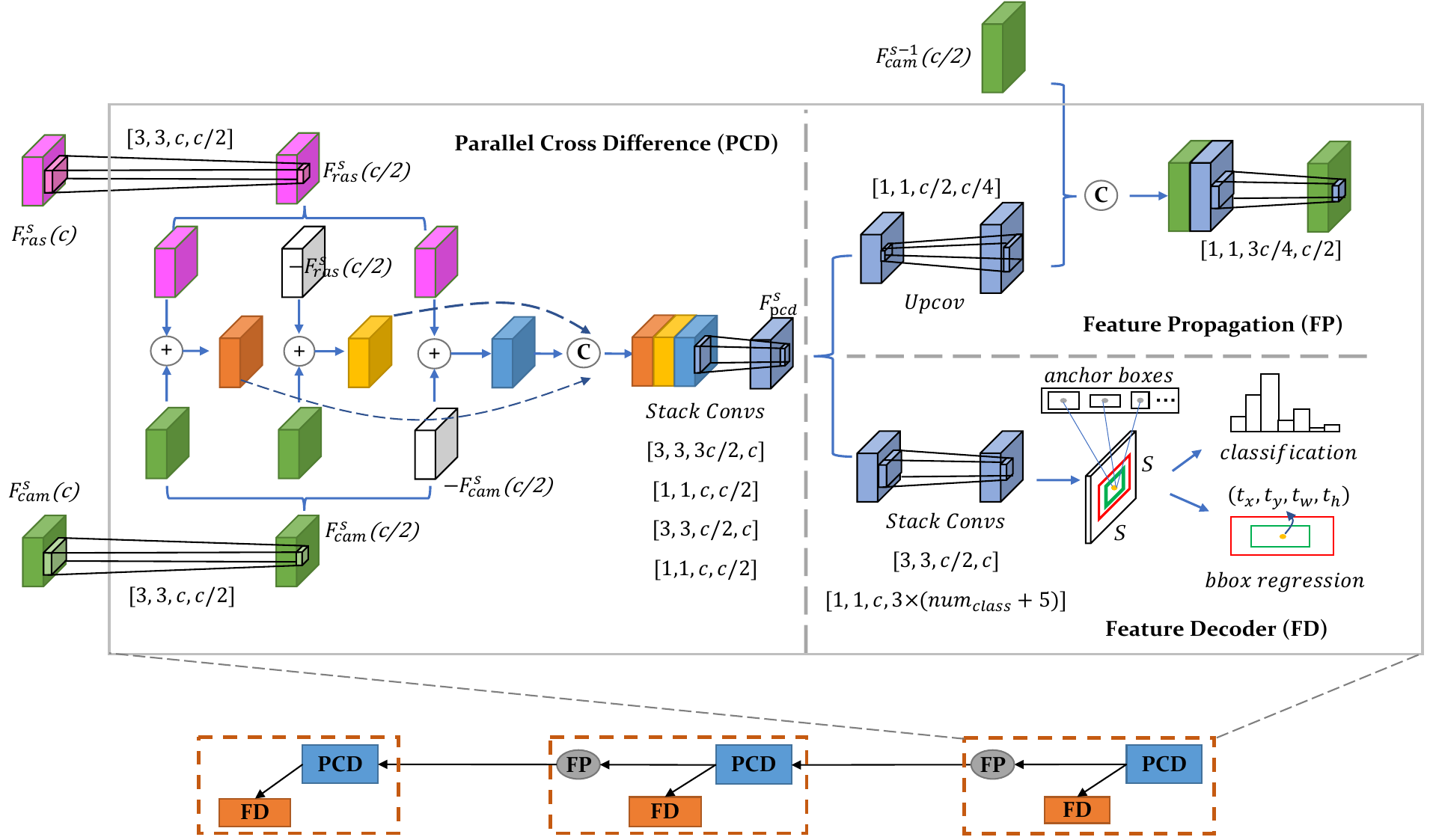}
	\end{center}
	\caption{The illustration of the detailed architecture of the parallel cross difference (PCD), feature propagation (FP) and feature decoder (FD) modules. The feature difference between rasterized and camera images calculated by the proposed PCD modules is one of the key designs of our work.
	$F^s_{cam}(c)$ represents the feature vector at $s$ level with $c$ channels.
	The $[3, 3, c, c/2]$ represents a $3 \times 3$ convolution.
	Its input vector has $c$ channels and its output has $c/2$ channels.
	}
	\label{fig:detailed-diff-net}
	\vspace{-0.2cm}
\end{figure*}

As we mentioned earlier, a conventional approach directly cross-compares object detection results against existing map elements to infer possible changes.
Note that it is not a trivial problem since missing or redundant elements and localization noises make them not necessarily a group of one-to-one comparisons in most cases.
Inspired by this process, we employ a deep neural network (DNN) that transforms comparisons in instance space to feature space, denoted as the parallel cross difference (PCD) network, as shown in Figure~\ref{fig:detailed-diff-net}.
Specifically, the PCD module calculates the difference between the two extracted features.
Its output features pass through 4 convolution layers and are then processed by a feature decoder (FD), finally producing change detection results.
This design leverages deep learning networks' powerful capabilities that they can generalize well in solving complicated problems.
Our experiments also demonstrate that the proposed method achieves better performance, as shown in Section~\ref{sec:experiment}.

Similar to YOLOv3~\cite{redmon2018yolov3}, we also predict bounding boxes at three different scales. As shown in Figures~\ref{fig:diff-net} and~\ref{fig:detailed-diff-net}, features outputted by the PCD module at a coarser scale go through a feature propagation (FP) module. They are upscaled to a finer scale and then concatenated with camera features in the finer scale. After another convolution, the resulted features are passed to the PCD module at a finer scale.

\subsection{Anchor-based Detection}
As mentioned earlier, the output features of the PCD module are processed by a feature decoder (FD) module, which produces final detection bounding boxes.
We first perform a $3\times3$ convolution in the FD module to lift the feature channel dimension from $c/2$ to $c$.
Then, a $1\times1$ convolution is applied to generate the region proposals, resulting in the final tensor with a channel number  $S\times S \times \left[ 3\times(num\_class + 5) \right] $, where $num\_class$ represents the number of the change categories (3 for $to\_add$, $to\_del$ and $correct$), $5$ represents the location and confidence of the bounding box, and $3$ denotes the number of anchor boxes in one of $S \times S$ grid cells ($S=7$).

Similar to YOLOv3~\cite{redmon2018yolov3}, we have two branches for change detection.
One outputs the change categories with softmax operations.
The other infers elements' geometric locations $t_x$, $t_y$, $t_w$, and $t_h$ with respect to necessary width and height priors $p_w$ and $p_h$ (See~\cite{redmon2018yolov3} for details).
Finally, the non-maximum suppression (NMS) method from~\cite{bodla2017soft} is used to eliminate redundant detections.

\subsection{Spatio-temporal Fusion}
\label{section:convlstm-net}

\begin{figure*}[htb]
	\begin{center}
		\includegraphics[width=0.8\linewidth]{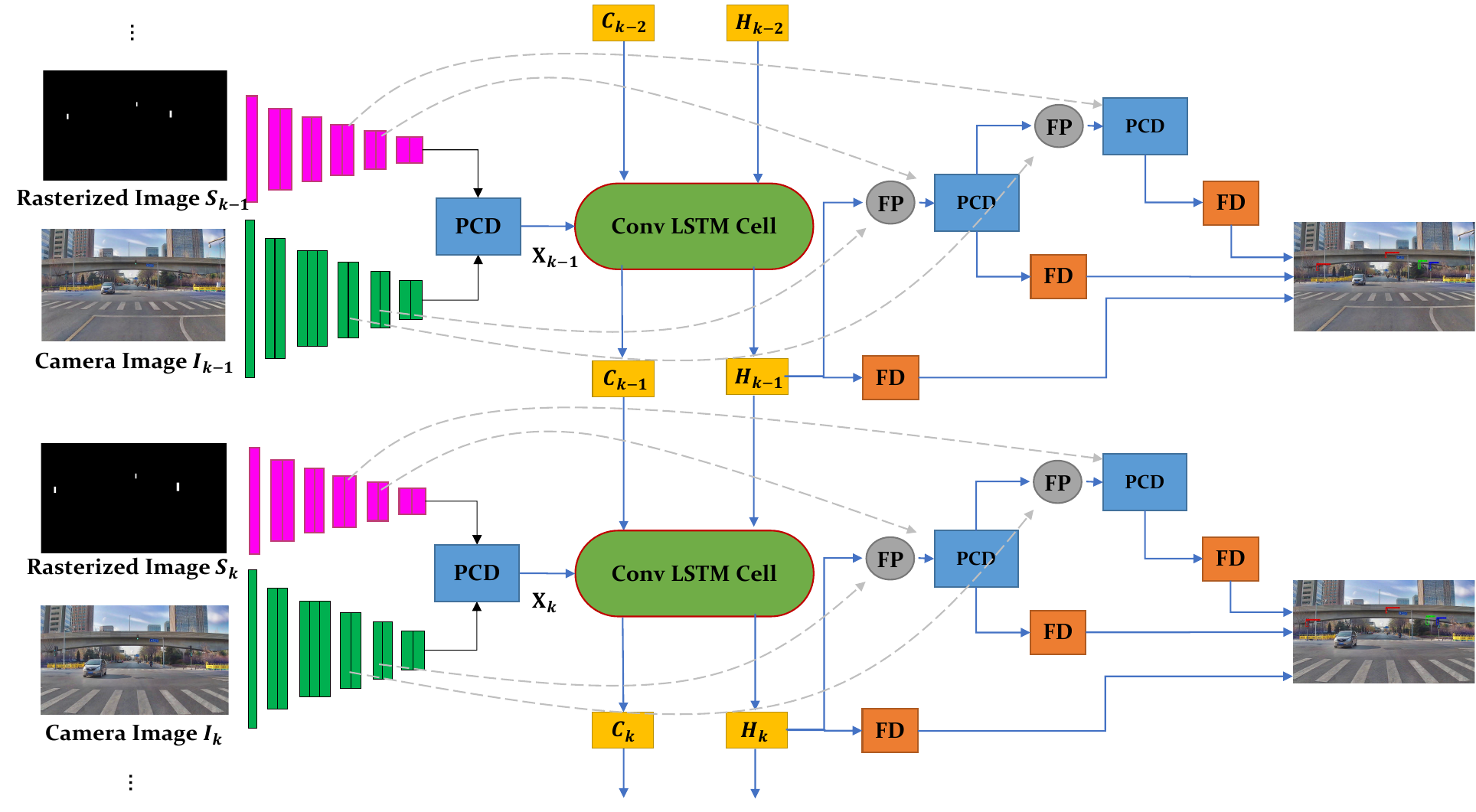}
	\end{center}
	\caption{The illustration of the network architecture with the ConvLSTM structure in the coarsest image resolution scale, which enables the temporal feature fusion between consecutive frames and improves the overall change detection performance.
	$C_k$ and $H_k$ denote the hidden state and the cell state in the ConvLSTM, respectively.}
	\label{fig:diff-net-lstm}
\end{figure*}

Essentially, similar to object detection in autonomous driving applications, the data is acquired as a video stream instead of sparse images, and detection results in the map change detection task are temporally correlated.
Therefore, inspired by~\cite{duzcceker2020deepvideomvs}, the ConvLSTM~\cite{shi2015convolutional} is incorporated to let features flow from history frames to the current time step, thus improving the overall change detection performance.
As shown in Figure~\ref{fig:diff-net-lstm}, let $\bm{X}_k$ denote the output of the PCD module at the k-th frame.
Our ConvLSTM lets latent temporal information flow in the coarsest image scale.
For finer scales, we apply skip connections that directly connect encoded features with corresponding FP or PCD modules yielding the same architecture illustrated in Figure~\ref{fig:diff-net}.
Similar to~\cite{shi2015convolutional}, both ELU activation~\cite{clevert2015fast} and layer normalization~\cite{ba2016layer} are adopted in our ConvLSTM implementation.

\subsection{Loss Function}
\label{section:loss-function}
The overall loss $L$ can be formulated as follows:

\begin{equation}\label{eq:fused_loss}
	Loss(\bm{D}, \widehat{\bm{D}}) = \lambda_{1} L_{GIoU} + \lambda_{2} L_{conf} + \lambda_{3} L_{prob}
\end{equation}
where $L_{GIoU}$ is the localization loss, $L_{conf}$ is the confidence loss, and $L_{prob}$ is the category probability loss. $\lambda_{1}, \lambda_{2}, \lambda_{3}$ are loss weights and are set as 1.0 in the experiments.

To boost inferred bounding boxes' location precision, especially for non-overlapping ones, we adopt the GIoU~\cite{rezatofighi2019generalized} loss as the localization metric, as defined in Eq~(\ref{eq:giou_loss}).

\begin{equation}\label{eq:giou_loss}
	\begin{aligned}
		L_{GIoU} = 1 -  \frac{1}{N}\sum_{i=1}^{N}( \frac{D_{i}\cap \widehat{D}_{i}}{D_{i}\cup \widehat{D}_{i}} -  \frac{f_{ch}(D_{i},\widehat{D}_{i})\setminus(D_{i}\cup \widehat{D}_{i})}{f_{ch}(D_{i},\widehat{D}_{i})}) \\
	\end{aligned}
\end{equation}
where $D_{i}$ denotes $i$-th bounding box in the HMCD results, $\widehat{D}_{i}$ represents the corresponding ground truth of $D_{i}$, $\cap$ computes the intersection area of two bounding boxes, $\cup$ computes the union area, $f_{ch}()$ computes the area of the minimum enclosing convex hull of a set of bounding boxes.

To improve the performance of complex, misclassified examples, we introduce a confidence loss that leverages the focal loss~\cite{lin2017focal}. It is defined as follows:

\begin{equation}\label{eq:conf_loss}
	\begin{aligned}
		\bm{L}_{conf}  = & \lambda_{obj}\sum_{i=0}^{S^{2}}\sum_{j=0}^{B}\mathbbm{1}_{ij}^{obj}-\alpha(\widehat{C}_{i}^{j} - C_{i}^{j})^{\gamma} f_{ce}(\widehat{C}_{i}^{j}, C_{i}^{j})+ \\
		& \lambda_{noobj}\sum_{i=0}^{S^{2}}\sum_{j=0}^{B} \mathbbm{1}_{ij}^{noobj} -(1-\alpha)(\widehat{C}_{i}^{j} - C_{i}^{j})^{\gamma} f_{ce}(\widehat{C}_{i}^{j}, C_{i}^{j}) \\
	\end{aligned}
\end{equation}
where $S^{2}$ is the number of the grid cells, $B$ is the number of the anchor boxes within a grid cell, $f_{ce}()$ represents the sigmoid cross entropy, $C_{i}^{j}$ represents the confidence score of the $j$-th bounding box in the $i$-th grid cell, $\widehat{C}_{i}^{j}$ represents the corresponding ground truth confidence values (1.0 if object exists and 0.0 if object doesn't exist), $\mathbbm{1}_{ij}^{obj}$ denotes
that the $j$-th bounding box predictor in cell $i$ is ``responsible"
for that prediction. For focal loss parameters $\alpha$ and $\gamma$, we set them as 0.5 and 2.0, respectively.

$\bm{L}_{prob} $ is the change category prediction loss, which is formulated as following:
\begin{equation}\label{eq:prob_loss}
	\bm{L}_{prob} = \sum_{i=0}^{S^{2}} \mathbbm{1}_{i}^{obj}\sum_{c\in classes}f_{ce}(\widehat{P}_{i}^{c}, P_{i}^{c})
\end{equation}
where $classes = \{correct, to\_del, to\_add\}$, $P_{i}^{c}$ represents the detection score of the $c$-th category in the $i$-th grid cell, and $\mathbbm{1}_{i}^{obj}$ denotes if the object appears in the $i$-th grid cell.

\section{Experiments}
\label{sec:experiment}

\subsection{Datasets}
To the best of our knowledge, there are no public datasets that focus on the HD map change detection task.
Therefore, we recruited our self-driving vehicles equipped with standard sensors, such as LiDARs, cameras, IMUs, and GNSS receivers.
We divide our data into three datasets, SICD, VSCD, and R-VSCD.
In SICD and VSCD datasets, we collected data in Yizhuang District, Beijing city, the capital city of one of the largest developing countries, where environmental changes, for example, road constructions, are common.
To fully validate different methods, we propose synthesizing change events as environmental changes are rare.
Furthermore, to meet different needs, the SICD dataset contains 205,076 isolated images while the VSCD contains 3,750 short video clips.
The R-VSCD dataset includes 44 video clips where actual traffic signal changes happened.
Furthermore, it is worth mentioning that the R-VSCD dataset includes data from four different cities in China, Beijing, Cangzhou, Guangzhou, and Shanghai.

\subsection{Performance Analysis of SICD/VSCD Dataset}
As we formulate our problem as a detection task, we adopt the mAP as our primary performance metric.
YOLOv3~\cite{redmon2018yolov3} + D is the conventional method that depends on a standard object detector together with necessary association and difference calculation algorithms.
When we calculate the mAP for $to\_del$ cases, YOLOv3 typically does not output bounding boxes in the area because there is no object.
Under such circumstances, we consider bounding boxes before the NMS step as candidates and take $1.0-c$ as the confidence score, where $c$ is the original confidence score in YOLOv3.
Diff-Net+ConvLSTM is our proposed method with the spatio-temporal fusion module.
In Table~\ref{performance_analysis}, we give a quantitative analysis of each method.
Note our vast performance improvement over the conventional method in both the SICD and VSCD datasets.
The end-to-end learning-based network achieves joint optimization of the change detection task, yielding significantly better performance overall.
In terms of video data, the ConvLSTM aided version performs even better and achieves 76.1\% mAP.

\begin{table}
	\centering
		\begin{tabular}{|l|c|c|}
			\hline
			\multirow{2}{*}{Method} & \multicolumn{2}{c|}{mAP} \\ \cline{2-3}
			& SICD & VSCD \\
			\hline\hline
			Yolov3~\cite{redmon2018yolov3}+D & 0.437 & 0.423 \\
			Diff-Net & \textbf{0.876} & 0.678 \\
			Diff-Net+ConvLSTM & - & \textbf{0.761} \\
			\hline
		\end{tabular}
	\caption{Comparison of the change detection performance using the mAP metric. Our wide improvement over the conventional method, Yolov3~\cite{redmon2018yolov3} + D, is notable. Meanwhile, the benefits of the ConvLSTM based spatio-temporal fusion module are visible.
	}
	\label{performance_analysis}
\end{table}

\begin{table}
	\begin{center}
		\begin{tabular}{|l|c|}
			\hline
			{Method} & {Top-1 Accuracy} \\ \cline{2-2}
			\hline\hline
			Yolov3~\cite{redmon2018yolov3}+D & 0.558 \\
			Diff-Net & 0.725  \\
			Diff-Net+ConvLSTM & $\bm{0.810}$  \\
			\hline
		\end{tabular}
	\end{center}
	\caption{Comparison of the change detection performance using the dataset with real HD map changes. The problem is formulated as a classification problem to determine the correct change category. Our proposed method outperforms the baseline method.}
	\label{HMCD_online_Quantitative}
	\vspace{-0.4cm}
\end{table}

\begin{figure*}[htb]
	\centering
	\includegraphics[width=1.0\linewidth]{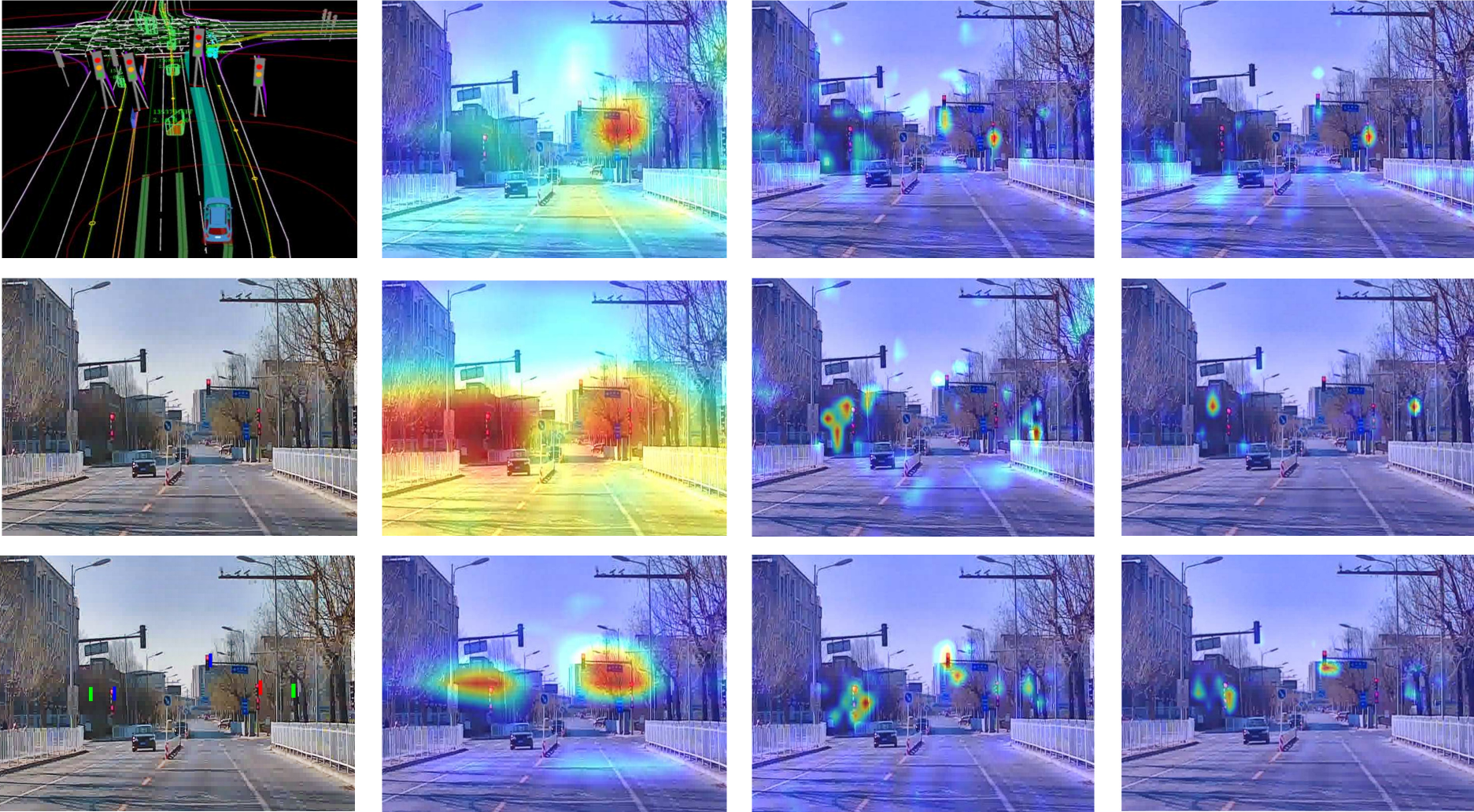}
	\caption{Visualization of three channels (from top to bottom) of the PCD module's output features. In the left-most column, from top to bottom, there are the HD map, a camera image, and a camera image with the ground truth of the change detection results. The blue, green, and red masks represent the $correct$, $to\_del$, and $to\_add$ change categories, respectively. For the other three columns, the features are visualized as heatmaps. From left to right, we show the features in different image scales in a coarse-to-fine pattern. Note that the features accurately spread in areas of interest in the images.}
	\label{fig:Visualization}
\end{figure*}

\subsection{Performance Analysis of R-VSCD Dataset}
As mentioned before, we introduced an R-VSCD dataset where actual traffic signal changes happened.
In this experiment, we evaluate the performance of the proposed methods in detecting HD map changes in the real world.
Since the number of changes in the R-VSCD dataset is too limited (HD map changes are rare) to produce a meaningful mAP value, we choose to evaluate the top-1 accuracy in this experiment.
It is known that there is zero or one change case in each video clip, so the problem becomes a classification problem to determine the correct change category of the video clip, $correct$, $to\_add$, or $to\_del$.
More specifically, we run our change detector for all frames in the video clip and determine the outcome using majority voting.
The top-1 accuracy results of each method are reported in Table~\ref{HMCD_online_Quantitative}.
The ConvLSTM aided version achieves 81.0\% top-1 accuracy and outperforms the baseline method.
It is worth mentioning that the R-VSCD includes data from 4 different cities.
However, our training data was collected in Beijing, China only.
It makes the problem sufficiently challenging since traffic signals look markedly different from city to city.

\subsection{Feature Visualization}
To help us interpret the effectiveness of the PCD module, we visualize three channels (from top to bottom) of the PCD's final output features $F^s_{pcd}$ in Figure~\ref{fig:Visualization}.
Note that the features accurately cover the areas of interest in the images.
Notably, no objects exist in the camera image for the $to\_del$ changes.
This implies that they are compelling features for the HD map change detection task.
Also, interestingly, we find that features in a coarser scale focus more on larger objects, while features in a more refined scale are for smaller ones.
This strictly follows the purpose of our design.

\section{Conclusion}
We have proposed a learning-based HD map change detection network designed for autonomous driving applications.
Instead of resolving the problem in several isolated steps, contrary to conventional methods, the proposed method constructs an end-to-end network that infers the map changes directly in a joint framework, yielding substantially more outstanding performance.
The excellent performance makes our method ready to be integrated into an HD map updating pipeline and support the regular operation of a self-driving fleet.
Three datasets have been introduced in this work to fully validate our method, including synthetic and natural HD map changes.
The datasets are to be released to the academic community soon.
Our future work includes expanding our methods for more map elements and handling irregularly shaped objects.

\bibliography{egbib}

\begin{thebibliography}{10}
\providecommand{\url}[1]{#1}
\csname url@samestyle\endcsname
\providecommand{\newblock}{\relax}
\providecommand{\bibinfo}[2]{#2}
\providecommand{\BIBentrySTDinterwordspacing}{\spaceskip=0pt\relax}
\providecommand{\BIBentryALTinterwordstretchfactor}{4}
\providecommand{\BIBentryALTinterwordspacing}{\spaceskip=\fontdimen2\font plus
\BIBentryALTinterwordstretchfactor\fontdimen3\font minus
  \fontdimen4\font\relax}
\providecommand{\BIBforeignlanguage}[2]{{%
\expandafter\ifx\csname l@#1\endcsname\relax
\typeout{** WARNING: IEEEtran.bst: No hyphenation pattern has been}%
\typeout{** loaded for the language `#1'. Using the pattern for}%
\typeout{** the default language instead.}%
\else
\language=\csname l@#1\endcsname
\fi
#2}}
\providecommand{\BIBdecl}{\relax}
\BIBdecl

\bibitem{pannen2019hd}
D.~Pannen, M.~Liebner, and W.~Burgard, ``Hd map change detection with a boosted
  particle filter,'' in \emph{Proceedings of the International Conference on
  Robotics and Automation (ICRA)}.\hskip 1em plus 0.5em minus 0.4em\relax IEEE,
  2019, pp. 2561--2567.

\bibitem{heo2020hd}
M.~Heo, J.~Kim, and S.~Kim, ``Hd map change detection with cross-domain deep
  metric learning,'' in \emph{Proceedings of the IEEE/RSJ International
  Conference on Intelligent Robots and Systems (IROS)}.\hskip 1em plus 0.5em
  minus 0.4em\relax IEEE, 2020, pp. 10\,218--10\,224.

\bibitem{radke2005image}
R.~J. Radke, S.~Andra, O.~Al-Kofahi, and B.~Roysam, ``Image change detection
  algorithms: A systematic survey,'' \emph{IEEE Transactions on Image
  Processing (TIP)}, vol.~14, no.~3, pp. 294--307, 2005.

\bibitem{taneja2015geometric}
A.~Taneja, L.~Ballan, and M.~Pollefeys, ``Geometric change detection in urban
  environments using images,'' \emph{IEEE Transactions on Pattern Analysis and
  Machine Intelligence (TPAMI)}, vol.~37, no.~11, pp. 2193--2206, 2015.

\bibitem{sakurada2015change}
K.~Sakurada and T.~Okatani, ``Change detection from a street image pair using
  cnn features and superpixel segmentation.'' in \emph{Proceedings of the
  British Machine Vision Conference (BMVC)}, vol.~61, 2015, pp. 1--12.

\bibitem{qin20163d}
R.~Qin, J.~Tian, and P.~Reinartz, ``{3D} change detection – approaches and
  applications,'' \emph{ISPRS Journal of Photogrammetry and Remote Sensing},
  vol. 122, pp. 41--56, 2016.

\bibitem{palazzolo2018fast}
E.~Palazzolo and C.~Stachniss, ``Fast image-based geometric change detection
  given a {3D} model,'' in \emph{Proceedings of the IEEE International
  Conference on Robotics and Automation (ICRA)}.\hskip 1em plus 0.5em minus
  0.4em\relax IEEE, 2018, pp. 6308--6315.

\bibitem{lei2020hierarchical}
Y.~Lei, D.~Peng, P.~Zhang, Q.~Ke, and H.~Li, ``Hierarchical paired channel
  fusion network for street scene change detection,'' \emph{IEEE Transactions
  on Image Processing (TIP)}, vol.~30, pp. 55--67, 2020.

\bibitem{golparvar2011monitoring}
M.~Golparvar-Fard, F.~Pena-Mora, and S.~Savarese, ``Monitoring changes of {3D}
  building elements from unordered photo collections,'' in \emph{Proceedings of
  the IEEE International Conference on Computer Vision Workshops
  (ICCVW)}.\hskip 1em plus 0.5em minus 0.4em\relax IEEE, 2011, pp. 249--256.

\bibitem{agarwal2011building}
S.~Agarwal, Y.~Furukawa, N.~Snavely, I.~Simon, B.~Curless, S.~M. Seitz, and
  R.~Szeliski, ``Building rome in a day,'' \emph{Communications of the ACM},
  vol.~54, no.~10, pp. 105--112, 2011.

\bibitem{furukawa2015multi}
Y.~Furukawa and C.~Hern{\'a}ndez, ``{Multi-View Stereo}: A tutorial,''
  \emph{Foundations and Trends{\textregistered} in Computer Graphics and
  Vision}, vol.~9, no. 1-2, pp. 1--148, 2015.

\bibitem{pollard2007change}
T.~Pollard and J.~L. Mundy, ``Change detection in a 3-d world,'' in
  \emph{Proceedings of the IEEE International Conference on Computer Vision and
  Pattern Recognition (CVPR)}.\hskip 1em plus 0.5em minus 0.4em\relax IEEE,
  2007, pp. 1--6.

\bibitem{ulusoy2014image}
A.~O. Ulusoy and J.~L. Mundy, ``Image-based 4-d reconstruction using 3-d change
  detection,'' in \emph{Proceedings of the European Conference on Computer
  Vision (ECCV)}.\hskip 1em plus 0.5em minus 0.4em\relax Springer, 2014, pp.
  31--45.

\bibitem{taneja2011image}
A.~Taneja, L.~Ballan, and M.~Pollefeys, ``Image based detection of geometric
  changes in urban environments,'' in \emph{Proceedings of the IEEE
  International Conference on Computer Vision (ICCV)}.\hskip 1em plus 0.5em
  minus 0.4em\relax IEEE, 2011, pp. 2336--2343.

\bibitem{qin20143d}
R.~Qin and A.~Gruen, ``{3D} change detection at street level using mobile laser
  scanning point clouds and terrestrial images,'' \emph{ISPRS Journal of
  Photogrammetry and Remote Sensing}, vol.~90, pp. 23--35, 2014.

\bibitem{sakurada2013detecting}
K.~Sakurada, T.~Okatani, and K.~Deguchi, ``Detecting changes in {3D} structure
  of a scene from multi-view imagescaptured by a vehicle-mounted camera,'' in
  \emph{Proceedings of the IEEE International Conference on Computer Vision and
  Pattern Recognition (CVPR)}.\hskip 1em plus 0.5em minus 0.4em\relax IEEE,
  2013, pp. 137--144.

\bibitem{eden2008using}
I.~Eden and D.~B. Cooper, ``Using {3D} line segments for robust and efficient
  change detection from multiple noisy images,'' in \emph{Proceedings of the
  European Conference on Computer Vision (ECCV)}.\hskip 1em plus 0.5em minus
  0.4em\relax Springer, 2008, pp. 172--185.

\bibitem{zhan2017change}
Y.~Zhan, K.~Fu, M.~Yan, X.~Sun, H.~Wang, and X.~Qiu, ``Change detection based
  on deep siamese convolutional network for optical aerial images,'' \emph{IEEE
  Geoscience and Remote Sensing Letters}, vol.~14, no.~10, pp. 1845--1849,
  2017.

\bibitem{guo2018learning}
E.~Guo, X.~Fu, J.~Zhu, M.~Deng, Y.~Liu, Q.~Zhu, and H.~Li, ``Learning to
  measure change: Fully convolutional siamese metric networks for scene change
  detection,'' \emph{arXiv preprint arXiv:1810.09111}, 2018.

\bibitem{alcantarilla2018street}
P.~F. Alcantarilla, S.~Stent, G.~Ros, R.~Arroyo, and R.~Gherardi, ``Street-view
  change detection with deconvolutional networks,'' \emph{Autonomous Robots},
  vol.~42, no.~7, pp. 1301--1322, 2018.

\bibitem{sakurada2020weakly}
K.~Sakurada, M.~Shibuya, and W.~Wang, ``Weakly supervised silhouette-based
  semantic scene change detection,'' in \emph{Proceedings of the IEEE
  International Conference on Robotics and Automation (ICRA)}.\hskip 1em plus
  0.5em minus 0.4em\relax IEEE, 2020, pp. 6861--6867.

\bibitem{ren2016faster}
S.~Ren, K.~He, R.~Girshick, and J.~Sun, ``Faster {R-CNN}: Towards real-time
  object detection with region proposal networks,'' \emph{IEEE Transactions on
  Pattern Analysis and Machine Intelligence (TPAMI)}, vol.~39, no.~6, pp.
  1137--1149, 2016.

\bibitem{lin2017feature}
T.-Y. Lin, P.~Doll{\'a}r, R.~Girshick, K.~He, B.~Hariharan, and S.~Belongie,
  ``Feature pyramid networks for object detection,'' in \emph{Proceedings of
  the IEEE International Conference on Computer Vision and Pattern Recognition
  (CVPR)}.\hskip 1em plus 0.5em minus 0.4em\relax IEEE, 2017, pp. 2117--2125.

\bibitem{he2017mask}
K.~He, G.~Gkioxari, P.~Doll{\'a}r, and R.~Girshick, ``Mask {R-CNN},'' in
  \emph{Proceedings of the IEEE International Conference on Computer Vision
  (ICCV)}.\hskip 1em plus 0.5em minus 0.4em\relax IEEE, 2017, pp. 2961--2969.

\bibitem{redmon2018yolov3}
J.~Redmon and A.~Farhadi, ``Yolov3: An incremental improvement,'' \emph{arXiv
  preprint arXiv:1804.02767}, 2018.

\bibitem{law2018cornernet}
H.~Law and J.~Deng, ``Cornernet: Detecting objects as paired keypoints,'' in
  \emph{Proceedings of the European Conference on Computer Vision
  (ECCV)}.\hskip 1em plus 0.5em minus 0.4em\relax Springer, 2018, pp. 734--750.

\bibitem{liu2016ssd}
W.~Liu, D.~Anguelov, D.~Erhan, C.~Szegedy, S.~Reed, C.-Y. Fu, and A.~C. Berg,
  ``{SSD}: Single shot multibox detector,'' in \emph{Proceedings of the
  European Conference on Computer Vision (ECCV)}.\hskip 1em plus 0.5em minus
  0.4em\relax Springer, 2016, pp. 21--37.

\bibitem{bansal2019chauffeurnet}
M.~Bansal, A.~Krizhevsky, and A.~Ogale, ``Chauffeurnet: Learning to drive by
  imitating the best and synthesizing the worst,'' in \emph{Proceedings of the
  Robotics: Science and Systems}, 2019.

\bibitem{buhler2020driving}
A.~B{\"u}hler, A.~Gaidon, A.~Cramariuc, R.~Ambrus, G.~Rosman, and W.~Burgard,
  ``Driving through ghosts: Behavioral cloning with false positives,''
  \emph{arXiv preprint arXiv:2008.12969}, 2020.

\bibitem{hecker2020learning}
S.~Hecker, D.~Dai, A.~Liniger, M.~Hahner, and L.~Van~Gool, ``Learning accurate
  and human-like driving using semantic maps and attention,'' in
  \emph{Proceedings of the IEEE/RSJ International Conference on Intelligent
  Robots and Systems (IROS)}.\hskip 1em plus 0.5em minus 0.4em\relax IEEE,
  2020, pp. 2346--2353.

\bibitem{casas2018intentnet}
S.~Casas, W.~Luo, and R.~Urtasun, ``Intentnet: Learning to predict intention
  from raw sensor data,'' in \emph{Proceedings of the Conference on Robot
  Learning}.\hskip 1em plus 0.5em minus 0.4em\relax PMLR, 2018, pp. 947--956.

\bibitem{chai20a}
Y.~Chai, B.~Sapp, M.~Bansal, and D.~Anguelov, ``Multipath: Multiple
  probabilistic anchor trajectory hypotheses for behavior prediction,'' in
  \emph{Proceedings of the Conference on Robot Learning}.\hskip 1em plus 0.5em
  minus 0.4em\relax PMLR, 2020, pp. 86--99.

\bibitem{Djuric_2020_WACV}
N.~Djuric, V.~Radosavljevic, H.~Cui, T.~Nguyen, F.-C. Chou, T.-H. Lin,
  N.~Singh, and J.~Schneider, ``Uncertainty-aware short-term motion prediction
  of traffic actors for autonomous driving,'' in \emph{Proceedings of the IEEE
  Winter Conference on Applications of Computer Vision (WACV)}.\hskip 1em plus
  0.5em minus 0.4em\relax IEEE, 2020, pp. 2095--2104.

\bibitem{bodla2017soft}
N.~Bodla, B.~Singh, R.~Chellappa, and L.~S. Davis, ``Soft-nms – improving
  object detection with one line of code,'' in \emph{Proceedings of the IEEE
  International Conference on Computer Vision (ICCV)}, 2017, pp. 5561--5569.

\bibitem{duzcceker2020deepvideomvs}
A.~D{\"u}z{\c{c}}eker, S.~Galliani, C.~Vogel, P.~Speciale, M.~Dusmanu, and
  M.~Pollefeys, ``{DeepVideoMVS}: Multi-view stereo on videowith recurrent
  spatio-temporal fusion,'' \emph{arXiv preprint arXiv:2012.02177}, 2020.

\bibitem{shi2015convolutional}
X.~Shi, Z.~Chen, H.~Wang, D.-Y. Yeung, W.-K. Wong, and W.-c. Woo,
  ``Convolutional lstm network: A machine learning approach for precipitation
  nowcasting,'' \emph{arXiv preprint arXiv:1506.04214}, 2015.

\bibitem{clevert2015fast}
D.-A. Clevert, T.~Unterthiner, and S.~Hochreiter, ``Fast and accurate deep
  network learning by exponential linear units (elus),'' \emph{arXiv preprint
  arXiv:1511.07289}, 2015.

\bibitem{ba2016layer}
J.~L. Ba, J.~R. Kiros, and G.~E. Hinton, ``Layer normalization,'' \emph{arXiv
  preprint arXiv:1607.06450}, 2016.

\bibitem{rezatofighi2019generalized}
H.~Rezatofighi, N.~Tsoi, J.~Gwak, A.~Sadeghian, I.~Reid, and S.~Savarese,
  ``Generalized intersection over union: A metric and a loss for bounding box
  regression,'' in \emph{Proceedings of the IEEE International Conference on
  Computer Vision and Pattern Recognition (CVPR)}.\hskip 1em plus 0.5em minus
  0.4em\relax IEEE, 2019, pp. 658--666.

\bibitem{lin2017focal}
T.-Y. Lin, P.~Goyal, R.~Girshick, K.~He, and P.~Doll{\'a}r, ``Focal loss for
  dense object detection,'' in \emph{Proceedings of the IEEE International
  Conference on Computer Vision (ICCV)}, 2017, pp. 2980--2988.

\end{thebibliography}
\end{document}